\documentclass[11pt,a4paper]{article}
\usepackage[hyperref]{naaclhlt2019}
\usepackage{times}
\usepackage{latexsym}

\usepackage{url}
\usepackage{enumitem}
\usepackage{graphicx}
\usepackage{paralist}
\usepackage{xfrac}
\usepackage{float}
\usepackage{amsfonts}
\usepackage{amsmath}
\usepackage{amsthm}
\usepackage{multirow}
\usepackage[para]{threeparttable}
\usepackage{xcolor}

\newtheorem{definition}{Definition}

\newcommand{\GRU}{\mathrm{GRU}}

\DeclareMathOperator{\maxpool}{maxpool}

\aclfinalcopy % Uncomment this line for the final submission
\def\aclpaperid{1093} %  Enter the acl Paper ID here

\title{HiGRU: Hierarchical Gated Recurrent Units for\\ Utterance-level Emotion Recognition}

\author{Wenxiang Jiao$^1$, Haiqin Yang$^{2,3}$, Irwin King$^1$, {\rm and} Michael R. Lyu$^1$ \\
  $^1$~Department of Computer Science and Engineering, \\
  The Chinese University of Hong Kong, HKSAR, China \\
  $^2$~Meitu and $^3$~The Hang Seng University of Hong Kong, HKSAR, China \\
  {\tt \{wxjiao,king,lyu\}@cse.cuhk.edu.hk, hqyang@ieee.org } \\}

\date{}

\begin{document}
\maketitle
\begin{abstract}
In this paper, we address three challenges in utterance-level emotion recognition in dialogue systems: (1) the same word can deliver different emotions in different contexts; (2) some emotions are rarely seen in general dialogues; (3) long-range contextual information is hard to be effectively captured.  We therefore propose a hierarchical Gated Recurrent Unit (HiGRU) framework with a lower-level GRU to model the word-level inputs and an upper-level GRU to capture the contexts of utterance-level embeddings.   Moreover, we promote the framework to two variants, HiGRU with individual features fusion (HiGRU-f) and HiGRU with self-attention and features fusion (HiGRU-sf), so that the word/utterance-level individual inputs and the long-range contextual information can be sufficiently utilized.  Experiments on three dialogue emotion datasets, IEMOCAP, Friends, and EmotionPush demonstrate that our proposed HiGRU models attain at least 8.7\%, 7.5\%, 6.0\% improvement over the state-of-the-art methods on each dataset, respectively.  Particularly, by utilizing only the textual feature in IEMOCAP, our HiGRU models gain at least 3.8\% improvement over the state-of-the-art conversational memory network (CMN) with the trimodal features of text, video, and audio.
\end{abstract}

\section{Introduction}
{Emotion recognition is a significant artificial intelligence research topic due to the promising potential of developing empathetic machines for people.  Emotion is a universal phenomena across different cultures and mainly consists of six basic types: anger, disgust, fear, happiness, sadness, and surprise~\cite{book/Ekman71,journal/Ekman92}.  
}
\begin{figure}[t]
\centering
\captionsetup{belowskip=-1em}
\includegraphics[width=0.46\textwidth]{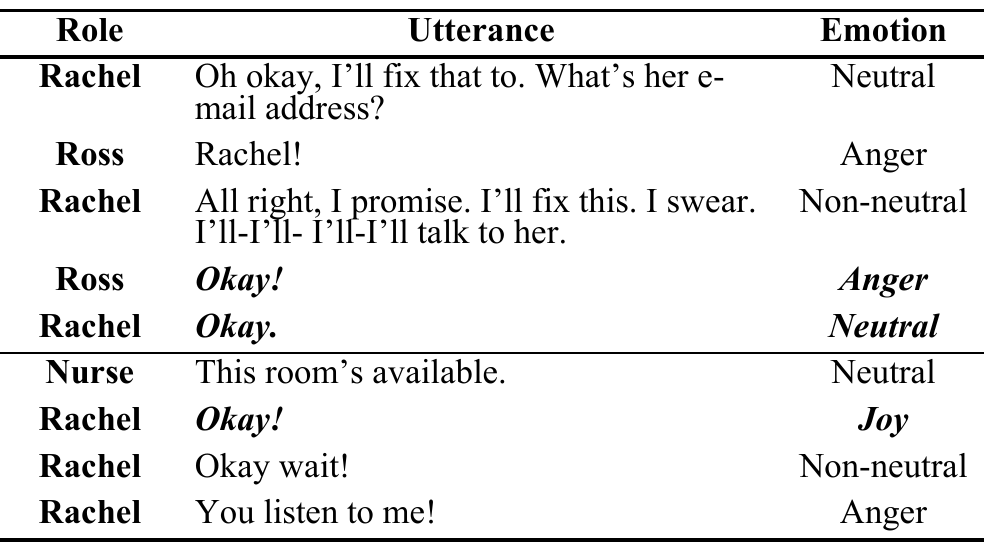}
\caption{\label{fig:friends} The word ``okay'' exhibits different emotions in the American television sitcom, Friends.}
\end{figure}

In this paper, we focus on textual dialogue systems because textual feature dominates the performance over audio and video features~\cite{DBLP:conf/emnlp/PoriaCG15,DBLP:conf/acl/PoriaCHMZM17}.  In utterance-level emotion recognition, an utterance~\cite{olson1977utterance} is a unit of speech bounded by breathes or pauses and its goal is to tag each utterance in a dialogue with the indicated emotion.

In this task, we address three challenges: First, the same word can deliver different emotions in different contexts.  For example, in Figure~\ref{fig:friends}, the word ``okay'' can deliver three different emotions, anger, neutral, and joy, respectively. Strong emotions like joy and anger may be indicated by the symbols ``!'' or ``?'' along the word. To identify a speaker's emotion precisely, we need to explore the dialogue context sufficiently.  Second, some emotions are rarely seen in general dialogues.  For example, people are usually calm and present a neutral emotion while only in some particular situations, they express strong emotions, like anger or fear.  Thus we need to be sensitive to the minority emotions while relieving the effect of the majority emotions.  Third, the long-range contextual information is hard to be effectively captured in an utterance/dialogue, especially when the length of an utterance/dialogue in the testing set is longer than those in the training set.

To tackle these challenges, we propose a hierarchical Gated Recurrent Unit (HiGRU) framework for the utterance-level emotion recognition in dialogue systems.  More specifically, HiGRU is composed by two levels of bidirectional GRUs, a lower-level GRU to model the word sequences of each utterance to produce individual utterance embeddings, and an upper-level GRU to capture the sequential and contextual relationship of utterances.  We further promote the proposed HiGRU to two variants, HiGRU with individual features fusion (HiGRU-f), and HiGRU with self-attention and features fusion (HiGRU-sf).
In HiGRU-f, the individual inputs, i.e., the word embeddings in the lower-level GRU and the individual utterance embeddings in the upper-level GRU, are concatenated with the hidden states to generate the contextual word/utterance embeddings, respectively.  In HiGRU-sf, a self-attention layer is placed on the hidden states from the GRU to learn long-range contextual embeddings, which are concatenated with the original individual embeddings and the hidden states to generate the contextual word/utterance embeddings.  Finally, the contextual utterance embedding is sent to a fully-connected (FC) layer to determine the corresponding emotion. To alleviate the effect of data imbalance issue, we follow~\cite{DBLP:conf/acl-socialnlp/Khosla18} to train our models by minimizing a weighted categorical cross-entropy. 

We summarize our contributions as follows:
\begin{compactitem}
    \item We propose a HiGRU framework to better learn both the individual utterance embeddings and the contextual information of utterances, so as to recognize the emotions more precisely.
    \item {We propose two progressive HiGRU variants, HiGRU-f and HiGRU-sf, to sufficiently incorporate the individual word/utterance-level information and the long-range contextual information respectively.} 
    \item We conduct extensive experiments on three textual dialogue emotion datasets, IEMOCAP, Friends, and EmotionPush. The results demonstrate that our proposed HiGRU models achieve at least 8.7\%, 7.5\%, 6.0\% improvement over state-of-the-art methods on each dataset, respectively.  Particularly, by utilizing only the textual feature in IEMOCAP, our proposed HiGRU models gain at least 3.8\% improvement over the existing best model, conversational memory network (CMN) with not only the text feature, but also the visual, and audio features.
\end{compactitem}

% Related work
\section{Related Work}
Text-based emotion recognition is a long-standing research topic~\cite{DBLP:conf/aaai/WilsonWH04,DBLP:conf/webi/YangLC07,Journal/MedhatHK14}. Nowadays, deep learning technologies have become dominant methods due to the outstanding performance.  Some prominent models include recursive autoencoders (RAEs)~\cite{DBLP:conf/emnlp/SocherPHNM11}, convolutional neural networks (CNNs)~\cite{DBLP:conf/emnlp/Kim14}, and recurrent neural networks (RNNs)~\cite{DBLP:conf/acl/Abdul-MageedU17}.
However, these models treat texts independently thus cannot capture the inter-dependence of utterances in dialogues~\cite{DBLP:conf/emnlp/Kim14,DBLP:conf/aaai/LaiXLZ15,DBLP:conf/eacl/GraveMJB17,DBLP:conf/emnlp/ChenSTLL16,DBLP:conf/naacl/YangYDHSH16}.  To exploit the contextual information of utterances, researchers mainly explore in two directions: (1) extracting contextual information among utterances, or (2) enriching the information embedded in the representations of words and utterances.

\paragraph{Contextual Information Extraction.}
The RNN architecture is a standard way to capture the sequential relationship of data.  \citeauthor{DBLP:conf/acl/PoriaCHMZM17} propose a bidirectional contextual long short-term memory (LSTM) network, termed bcLSTM, to model the context of textual features extracted by CNNs. \citeauthor{DBLP:conf/naacl/HazarikaPZCMZ18} improve bcLSTM by a conversational memory network (CMN) to capture the self and inter-speaker emotional influence, where GRU is utilized to model the self-influence and the attention mechanism is employed to excavate the inter-speaker emotional influence.  Though CMN is reported to attain better performance than bcLSTM on IEMOCAP~\cite{DBLP:conf/naacl/HazarikaPZCMZ18}, the memory network is too complicated for small-size dialogue datasets.

\paragraph{Representation Enrichment.}
Multimodal features have been utilized to enrich the representation of utterances~\cite{DBLP:conf/emnlp/PoriaCG15,DBLP:conf/acl/PoriaCHMZM17}.  Previous work indicate that textual features dominate the performance of recognizing emotions in contrast to visual or audio features~\cite{DBLP:conf/emnlp/PoriaCG15,DBLP:conf/acl/PoriaCHMZM17}.  Recently, the textual features are mainly extracted by CNNs to learn individual utterance embeddings~\cite{DBLP:conf/emnlp/PoriaCG15,DBLP:conf/acl/PoriaCHMZM17,DBLP:conf/aaai/ZahiriC18,DBLP:conf/naacl/HazarikaPZCMZ18}.  However, CNNs do not capture the contextual information within each utterance well. 
\begin{figure*}[htp]
\centering
\captionsetup{belowskip=-1em}
\includegraphics[width=0.96\textwidth]{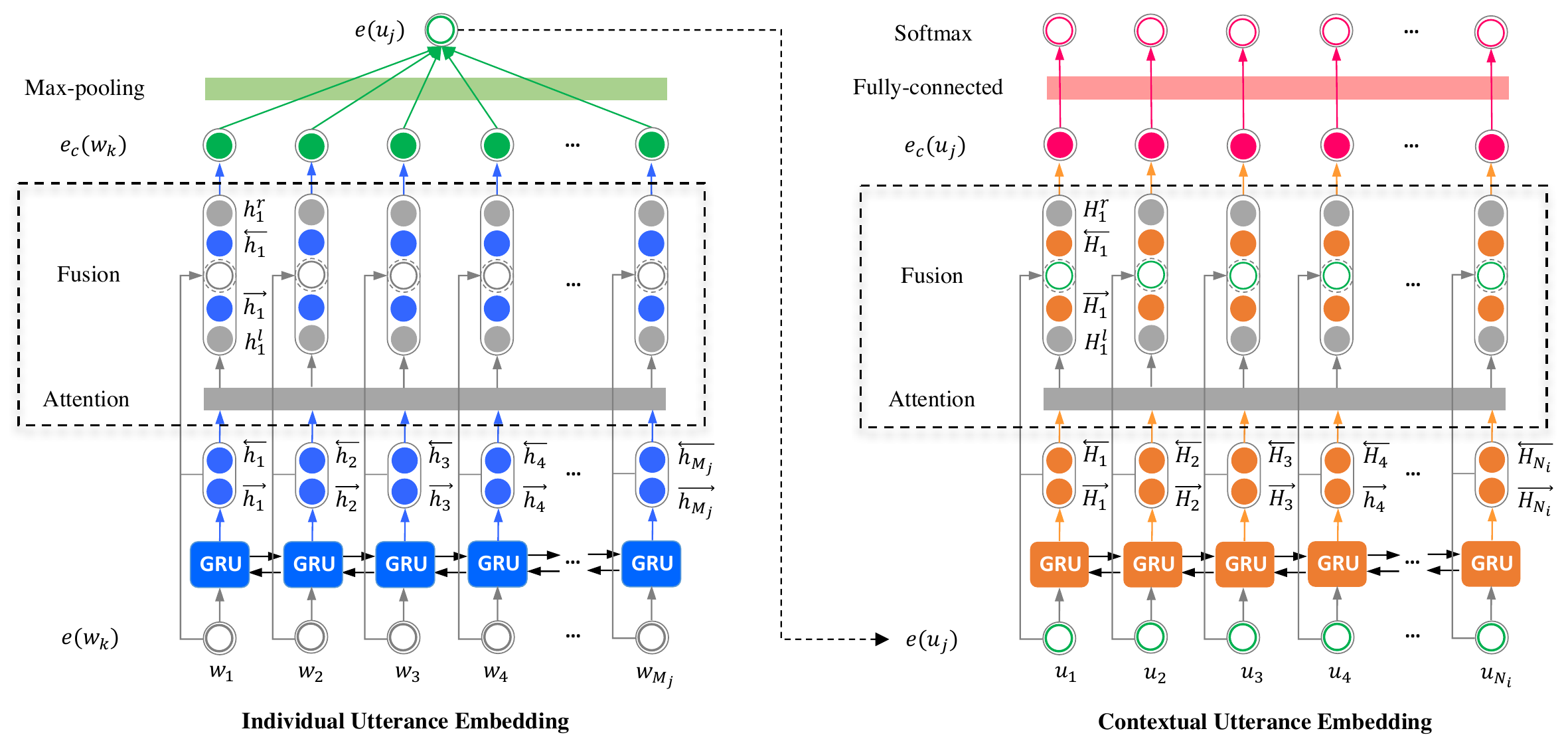}
\caption{\label{fig:architecture}The architecture of our proposed HiGRU-sf. ``Attention'' denotes self-attention. By removing the ``Attention" layer, we attain HiGRU-f, and by further removing the ``Fusion'' layer, we can recover the vanilla HiGRU.}
\end{figure*}

On the other hand, hierarchical RNNs have been proposed and demonstrated good performance in conventional text classification task~\cite{DBLP:conf/emnlp/TangQL15}, dialogue act classification~\cite{DBLP:conf/emnlp/LiuHTL17,DBLP:conf/aaai/KumarADJ18}, and speaker change detection~\cite{DBLP:conf/cikm/MengMJ17}. But they are not well explored in the task of utterance-level emotion recognition in dialogue systems.

\section{Approach}

The task of utterance-level emotion recognition is defined as follows:
\begin{definition}[\bf Utterance-level Emotion Recognition] Suppose we are given a set of dialogues, ${\cal D}=\{D_i\}_{i=1}^L$, where $L$ is the number of dialogues.  In each dialogue, $D_i=\{(u_j, s_j, c_j)\}_{j=1}^{N_i}$, is a sequence of $N_i$ utterances, where the utterance $u_j$ is spoken by the speaker $s_j\in {\cal S}$ with a certain emotion $c_j\in {\cal C}$.  All speakers compose the set ${\cal S}$ and the set ${\cal C}$ consists of all emotions, such as anger, joy, sadness, and neutral.  Our goal is to train a model ${\cal M}$ to tag each new utterance with an emotion label from ${\cal C}$ as accurately as possible.
\end{definition}

To solve this task, we propose a hierarchical Gated Recurrent Units (HiGRU) framework and extend two progressive variants, HiGRU with individual features fusion (HiGRU-f) and HiGRU with self-attention and features fusion (HiGRU-sf) (illustrated in Figure~\ref{fig:architecture}).

%\subsection
\subsection{HiGRU: Hierarchical GRU}

The vanilla HiGRU consists of two-level GRUs: the lower-level bidirectional GRU is to learn the individual utterance embedding by modeling the word sequence within an utterance and the upper-level bidirectional GRU is to learn the contextual utterance embedding by modeling the utterance sequence within a dialogue.

\paragraph{Individual Utterance Embedding.}
For the $j^{th}$ utterance in $D_i$, $u_j=\{w_k\}_{k=1}^{M_j}$, where $M_j$ is the number of words in the utterance $u_j$. {The corresponding sequence of individual word embeddings $\{ e(w_k)\}_{k=1}^{M_j}$ are fed into the lower-level bidirectional GRU~\cite{DBLP:conf/emnlp/ChoMGBBSB14} to learn the individual utterance embedding} in two opposite directions:
\begin{align}
\!\!\overrightarrow{h_k} = {\GRU}(e(w_k), \overrightarrow{h_{k-1}}),\\~~
\overleftarrow{h_k} = {\GRU}(e(w_k), \overleftarrow{h_{k+1}}) .
\end{align}
The two hidden states $\overrightarrow{h_k}$ and $\overleftarrow{h_k}$ are concatenated into $hs=[\overrightarrow{h_k}; \overleftarrow{h_k}]$ to produce the {\em contextual word embedding} for $w_k$ via the $\tanh$ activation function on a linear transformation:
\begin{align}
e_c(w_k) = \tanh(W_w\cdot hs + b_w) ,
\end{align}
where $W_w\in \mathbb{R}^{d_1\times 2d_1}$ and $\ b_w\in \mathbb{R}^{d_1}$ are the model parameters, $d_0$ and $d_1$ are the dimensions of word embeddings and the hidden states of the lower-level GRU, respectively.  

The individual utterance embedding is then obtained by max-pooling on the {contextual} word embeddings within the utterance:
\begin{align}
e(u_j) = \maxpool\left(\{ e_c(w_k)\}_{k=1}^{M_j}\right) .
\end{align}

\paragraph{Contextual Utterance Embedding.}
For the $i^{th}$ dialogue, $D_i=\{(u_j, s_j, c_j)\}_{j=1}^{N_i}$, the learned individual utterance embeddings, $\{ e(u_j)\}_{j=1}^{N_i}$, are fed into the upper-level bidirectional GRU to capture the sequential and contextual relationship of utterances in a dialogue:
\begin{align}
\overrightarrow{H_j} &= {\GRU}(e(u_j), \overrightarrow{H_{j-1}}),\\
\overleftarrow{H_j} &= {\GRU}(e(u_j), \overleftarrow{H_{j+1}}).
\end{align}
Here, {the hidden states of the upper-level GRU} are represented by $H_j\in \mathbb{R}^{d_2}$, to distinguish from those learned in the lower-level GRU denoted by $h_k$.  Accordingly, we can obtain the {\em contextual utterance embedding} by
\begin{align}
e_c(u_j) = \tanh(W_u\cdot Hs + b_u) ,
\end{align}
where $Hs=[\overrightarrow{H_j}; \overleftarrow{H_j}]$, $W_u\in \mathbb{R}^{d_2\times 2d_2}$ and $b_u\in \mathbb{R}^{d_2}$ are the model parameters, $d_2$ is the dimension of the hidden states in the upper-level GRU. Since the emotions are recognized at utterance-level, the learned contextual utterance embedding $e_c(u_j)$ is directly fed to a FC layer followed by a softmax function to determine the corresponding emotion label:
\begin{align}\label{eq:softmax}
\hat{y}_j = \textrm{softmax}(W_{fc}\cdot e_c(u_j) + b_{fc}),
\end{align}
where $\hat{y}_j$ is the predicted vector over all emotions, and $W_{fc}\in \mathbb{R}^{|\mathcal{C}|\times d_2}$, $\ b_{fc}\in \mathbb{R}^{|\mathcal{C}|}$.

\subsection{HiGRU-f: HiGRU + Individual Features Fusion}

The vanilla HiGRU contains two main issues: (1) the individual word/utterance embeddings are diluted with the stacking of layers; (2) the upper-level GRU tends to gather more contextual information from the majority emotions, which {deteriorates the overall model performance}.

To resolve these two problems, we propose to fuse individual word/utterance embeddings with the hidden states from GRUs so as to strengthen the information of each word/utterance in its contextual embedding. This variant is named as HiGRU-f, representing HiGRU with individual features fusion. Hence, the lower-level GRU can maintain individual word embeddings and the upper-level GRU can relieve the effect of majority emotions and attain a more precise utterance representation for different emotions. Specifically, the contextual embeddings are updated as:
\begin{align}
e_c(w_k) &= \tanh(W_w\cdot {hs}^{f} + b_w) ,\\
e_c(u_j) &= \tanh(W_u\cdot {Hs}^{f} + b_u),
\end{align}
where $W_w\in \mathbb{R}^{d_1\times(d_0+2d_1)}$, $W_u\in \mathbb{R}^{d_2\times(d_1+2d_2)}$,  ${hs}^{f}=[\overrightarrow{h_k}; e(w_k); \overleftarrow{h_k}]$, and ${Hs}^f=[\overrightarrow{H_j}; e(u_j); \overleftarrow{H_j}]$.

\subsection{HiGRU-sf: HiGRU + Self-Attention and Feature Fusion}

Another challenging issue is to extract the contextual information of long sequences, especially the sequences in the testing set that are longer than those in the training set~\cite{DBLP:journals/corr/BahdanauCB14}.  To fully utilize the global contextual information, we place a self-attention layer upon the hidden states of HiGRU and fuse the attention outputs with the individual word/utterance embeddings and the hidden states to learn the contextual word/utterance embeddings.  Hence, this variant is termed HiGRU-sf, representing HiGRU with self-attention and features fusion.

Particularly, we apply self-attention upon the forward and backward hidden states separately to produce the left context embedding, $h_k^l$ ($H_j^l$), and the right context embedding, $h_k^r$ ($H_j^r$), respectively.  This allows us to gather the unique global contextual information at the current step in two opposite directions and yield the corresponding contextual embeddings computed as follows:
\begin{align}
e_c(w_k) &= \tanh(W_w\cdot {hs}^{sf} + b_w) ,\\
e_c(u_j) &= \tanh(W_u\cdot {Hs}^{sf} + b_u) ,
\end{align}
where $W_w\in \mathbb{R}^{d_1\times(d_0+4d_1)}$, $W_u\in \mathbb{R}^{d_2\times(d_1+4d_2)}$, ${hs}^{sf}=[h_k^l;\overrightarrow{h_k}; e(w_k); \overleftarrow{h_k}; h_k^r]$, and ${Hs}^{sf}=[H_j^l;\overrightarrow{H_j}; e(u_j); \overleftarrow{H_j};H_j^r]$.

%\iffalse
% Self-attention
\begin{figure}[t]
\centering
\captionsetup{belowskip=-1em}
\includegraphics[width=0.4\textwidth]{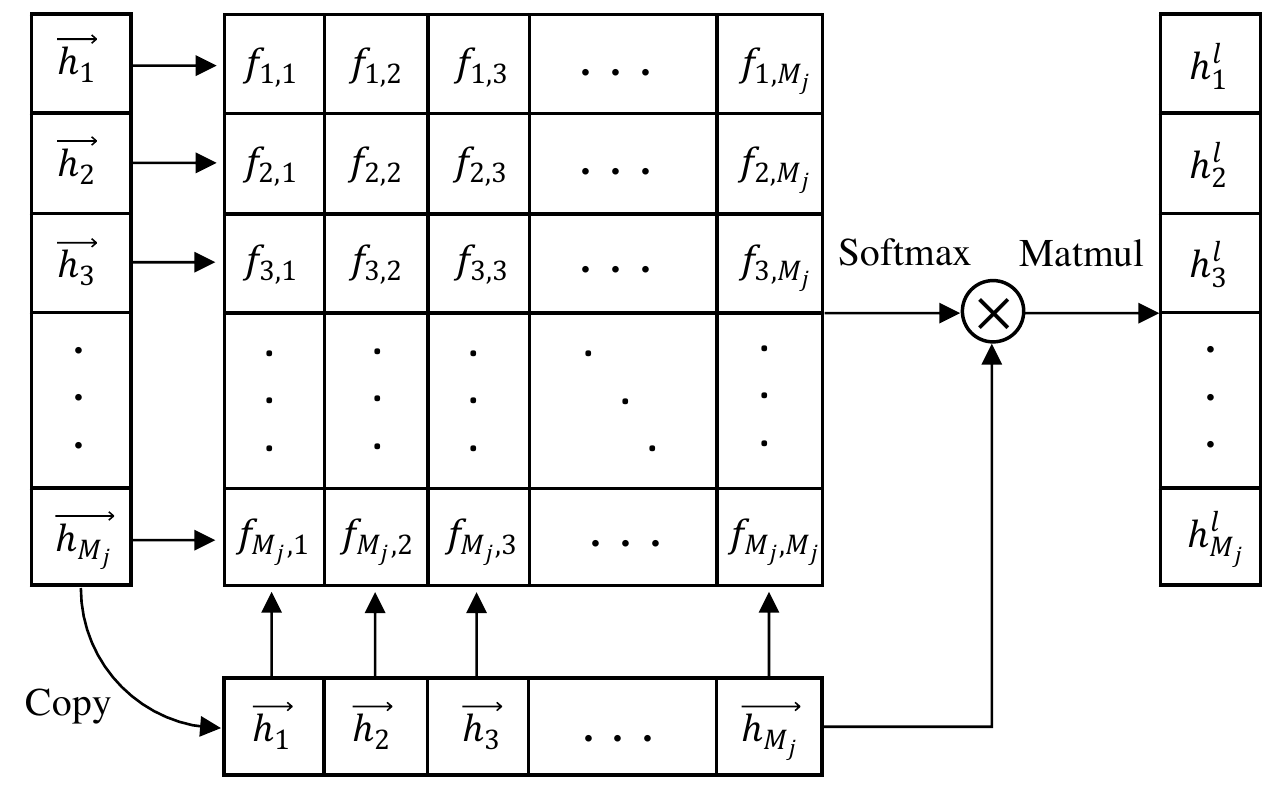}
\caption{\label{fig:selfatt} Self-attention over the forward hidden states of GRU.}
\end{figure}
%\fi

\paragraph{Self-Attention (SA).}
The self-attention mechanism is an effective non-recurrent architecture to compute the relation between one input to all other inputs and has {been} successfully applied in various natural language processing applications such as reading comprehension~\cite{DBLP:conf/ijcai/HuPHQW018}, and neural machine translation~\cite{DBLP:conf/nips/VaswaniSPUJGKP17}.  Figure~\ref{fig:selfatt} shows the dot-product SA over the forward hidden states of GRU to learn the left context $h_k^l$.
{Each element in the attention matrix is computed by}
\begin{eqnarray}
f(\overrightarrow{h_k}, \overrightarrow{h_p})=\left.%\lbrace
\begin{cases}%{lr}
\overrightarrow{h_k}^\top \overrightarrow{h_p},\quad &\mathrm{if}\ k,p\leq M_j, \\
-\infty,\quad &\mathrm{otherwise}.
\end{cases}\right.
\end{eqnarray}
An attention mask is then applied to waive the inner attention between the sequence inputs and paddings.  At each step, the corresponding left context $h_k^l$ is then computed by the weighted sum of all the forward hidden states: 
\begin{align}
\!\!h_k^l\! &\!=\!\!\sum_{p=1}^{M_j} a_{kp}\overrightarrow{h_p},\ 
a_{kp}= \frac{\scriptstyle\exp(f(\overrightarrow{h_k},\overrightarrow{h_p}))}{\scriptstyle\sum_{p'=1}^{M_j} \exp\left(f(\overrightarrow{h_k}, \overrightarrow{h_{p'}})\right)},
\end{align}
where $a_{kp}$ is the weight of $\overrightarrow{h_p}$ to be included in $h_k^l$. The right context $h_k^r$ can be computed similarly.

% Statistics of datasets
%\iffalse
\begin{table*}[t]
\small
\centering
\captionsetup{belowskip=-1em}
\begin{threeparttable}
\begin{tabular}{|l|ccc|ccccc|}
\hline
\multirow{2}{*}{\bf Dataset}
& \multicolumn{3}{c|}{\bf \#Dialogue (\#Utterance)}  
& \multicolumn{5}{c|}{\bf \#Emotion} \\
\cline{2-9}
& Train & Val & Test & Ang & Hap/Joy & Sad & Neu & Others  \\
\hline
IEMOCAP & 96 (3,569)  & 24 (721) & 31 (1,208) & 1,090 & 1,627 & 1,077 & 1,704 & 0 \\
Friends & 720 (10,561) & 80 (1,178) & 200 (2,764) & 759 & 1,710 & 498 & 6,530 & 5,006 \\
EmotionPush & 720 (10,733) & 80 (1,202) & 200 (2,807) & 140 & 2,100 & 514 & 9,855 & 2,133 \\
\hline
\end{tabular}
\end{threeparttable}
\caption{\label{table:statistics}Statistics of the textual dialogue datasets.}
\end{table*}

% Loss Function
\subsection{Model Training}

Following~\cite{DBLP:conf/acl-socialnlp/Khosla18} which attains the best performance
in the EmotionX shared task~\cite{DBLP:conf/acl-socialnlp/HsuK18}, we
minimize a weighted categorical cross-entropy on each utterance
of all dialogues to optimize the model parameters:
\begin{align}
loss = -\frac{1}{\sum_{i=1}^{L}N_i}\sum_{i=1}^{L} \sum_{j=1}^{N_i} \omega(c_j) \sum_{c=1}^{|\mathcal{C}|} y_j^c\log_2(\hat{y}_j^c),
\end{align}
where $y_j$ is the original one-hot vector of the emotion labels, and $y_j^c$ and $\hat{y}_j^c$ are the elements of $y_j$ and $\hat{y}_j$ corresponding to the class $c$.

Similar to~\cite{DBLP:conf/acl-socialnlp/Khosla18}, we assign the loss weight $\omega(c_j)$ inversely proportional to the number of training utterances in the class $c_j$, denoted by $I_c$, {i.e.}, assigning larger loss weights for the minority classes to relieve the data imbalance issue. The difference is that we add a constant $\alpha$ to adjust the smoothness of the distribution. Then, we have:
\begin{align}
\frac{1}{\omega(c)} = \frac{I_c^\alpha}{\sum_{c'=1}^{|\mathcal{C}|} I_{c'}^\alpha} .
\end{align}

\section{Experiments}

We conduct systematical experiments to demonstrate the advantages of our proposed HiGRU models.  

\subsection{Datasets}  
\label{ssec:dataset}

The experiments are carried out on three textual dialogue emotion datasets (see the statistics in Table~\ref{table:statistics}):

\paragraph{IEMOCAP\footnote{\url{https://sail.usc.edu/iemocap/}}.}
It contains approximately 12 hours of audiovisual data, including video, speech, motion capture of face, text transcriptions.  Following~\cite{DBLP:conf/acl/PoriaCHMZM17,DBLP:conf/naacl/HazarikaPZCMZ18}: (1) We apply the first four sessions for training and the last session for test; (2) The validation set is extracted from the shuffled training set with the ratio of 80:20; (3) We only evaluate the performance on four emotions: anger, happiness, sadness, neutral, and remove the rest utterances.

\paragraph{Friends\footnote{\url{http://doraemon.iis.sinica.edu.tw/emotionlines}}.}
The dataset is annotated from the Friends TV Scripts~\cite{DBLP:conf/acl-socialnlp/HsuK18}, where each dialogue in the dataset consists of a scene of multiple speakers.  Totally, there are 1,000 dialogues, which are {split} into 720, 80, and 200 dialogues for training, validation, and testing, respectively.  Each utterance in a dialogue is labeled by one of the eight emotions: anger, joy, sadness, neutral, surprise, disgust, fear, and non-neutral.

\paragraph{EmotionPush\footnote{\url{http://doraemon.iis.sinica.edu.tw/emotionlines}}.}
The dataset consists of private conversations between friends on the Facebook messenger collected by an App called EmotionPush, which is released for the EmotionX shared task~\cite{DBLP:conf/acl-socialnlp/HsuK18}.  Totally, there are 1,000 dialogues, which are {split} into 720, 80, 200 dialogue for training, validation, and testing, respectively.  All the utterances are categorized into one of the eight emotions as in the Friends dataset.

Following the setup of~\cite{DBLP:conf/acl-socialnlp/HsuK18}, in Friends and EmotionPush, we only evaluate the model performance on four emotions: anger, joy, sadness, and neutral, and we exclude the contribution of the rest emotion classes during training by setting their loss weights to zero.

\paragraph{Data Preprocessing.}
We preprocess the datasets by the following steps: (1) The utterances are split into tokens with each word being made into the lowercase; (2) All  non-alphanumerics except ``?'' and ``!'' are removed because these two symbols usually exhibit strong emotions, such as surprise, joy and anger; (3) We build a dictionary based on the words and symbols extracted, and follow ~\cite{DBLP:conf/acl/PoriaCHMZM17} to represent the tokens by the publicly available 300-dimensional word2vec\footnote{\url{https://code.google.com/archive/p/word2vec/}} vectors trained on 100 billion words from Google News. The tokens not included in the word2vec dictionary are initialized by randomly-generated vectors.

\subsection{Evaluation Metrics}

To conduct fair comparison, we adopt two metrics as~\cite{DBLP:conf/acl-socialnlp/HsuK18}, the weighted accuracy (WA) and unweighted accuracy (UWA):
\begin{align}
\mathrm{WA} = \sum_{c=1}^{|\mathcal{C}|} p_c\cdot a_c ,\quad \mathrm{UWA} = \frac{1}{|\mathcal{C}|}\sum_{c=1}^{|\mathcal{C}|} a_c ,
\end{align}
where $p_c$ is the percentage of the class $c$ in the testing set, and $a_c$ is the corresponding accuracy.

{Generally, recognizing strong emotions may provide more value than detecting the neutral emotion~\cite{DBLP:conf/acl-socialnlp/HsuK18}.  Thus, in Friends and EmotionPush, UWA is a more favorite evaluation metric because WA is heavily compromised with the large proportion of the neutral emotion.}

\subsection{Compared Methods}

Our proposed vanilla HiGRU, HiGRU-f, and HiGRU-sf\footnote{\small\url{https://github.com/wxjiao/HiGRUs}} are compared with the following state-of-the-art baselines:

\textbf{bcLSTM~\cite{DBLP:conf/acl/PoriaCHMZM17}}: a bidirectional contextual LSTM with multimodal features extracted by CNNs;

\textbf{CMN~\cite{DBLP:conf/naacl/HazarikaPZCMZ18}}: a conversational memory network with multimodal features extracted by CNNs;

\textbf{SA-BiLSTM~\cite{DBLP:conf/acl-socialnlp/LuoYC18}}: a self-attentive bidirectional LSTM model, a neat model achieving the second place of EmotionX Challenge~\cite{DBLP:conf/acl-socialnlp/HsuK18};

\textbf{CNN-DCNN~\cite{DBLP:conf/acl-socialnlp/Khosla18}}: a convolutional-deconvolutional autoencoder with more handmade features, the winner of EmotionX Challenge~\cite{DBLP:conf/acl-socialnlp/HsuK18};

\textbf{bcLSTM$_*$} and \textbf{bcGRU}: our implemented bcLSTM and bcGRU with the weighted loss on the textual feature extracted from CNNs.

\subsection{Training Procedure}

All our implementations are coded on the Pytorch framework.  To prevent the models fitting the order of data, we randomly shuffle the training set at the beginning of every epoch.

\paragraph{Parameters.}  
For \textbf{bcLSTM$_*$} and \textbf{bcGRU}, the CNN layer follows the setup of~\cite{DBLP:conf/emnlp/Kim14}, i.e., consisting of the kernels of 3, 4, and 5 with 100 feature maps each.  The convolution results of each kernel are fed to a max-over-time pooling operation.  The dimension of the hidden states of the upper-level bidirectional LSTM or GRU is set to 300.  For HiGRU, HiGRU-f, and HiGRU-sf, the dimensions of hidden states are set to 300 for both levels.  The final FC layer contains two sub-layers with 100 neurons each.

\paragraph{Training.}  
We adopt Adam~\cite{DBLP:journals/corr/KingmaB14} as the optimizer and set an initial learning rate,  $1\times10^{-4}$ for IEMOCAP and $2.5\times10^{-4}$ for Friends and EmotionPush, respectively.  An annealing strategy is utilized by decaying the learning rate by half every 20 epochs. Early stopping with a patience of 10 is adopted to terminate training based on the accuracy of the validation set.  Specifically, following the best models on each dataset, the parameters are tuned to optimize WA on the validation set of IEMOCAP and to optimize UWA on the validation set of Friends and EmotionPush, respectively. Gradient clipping with a norm of 5 is applied to model parameters. To prevent overfitting, dropout with a rate of 0.5 is applied after the contextual word/utterance embeddings, and the FC layer.

\paragraph{Loss weights.} 
For Friends and EmotionPush, as mentioned in Section~\ref{ssec:dataset}, the loss weights are set to zero except the four considered emotions, to ignore the others during training. Besides, the power rate $\alpha$ of loss weights is tested from 0 to 1.5 with a step of 0.25, and we use the best one for each model and dataset.

% Experimental results
% IEMOCAP
\begin{table}[t]
\centering
%\captionsetup{belowskip=-0.5em}
\begin{threeparttable}
\resizebox{\columnwidth}{!}{
\begin{tabular}{|l|cccc|cc|}
\hline
 \bf Model (Feat) & \bf Ang & \bf Hap & \bf Sad & \bf Neu & \bf WA & \bf UWA \\
\hline
\hline
bcLSTM\tnote{1} ~(T) & 76.07 & 78.97 & 76.23 & 67.44 & 73.6 & \underline{74.6} \\
(T+V+A) & 77.98 & 79.31 & 78.30 & 69.92 & 76.1 & \underline{76.3} \\
\hline
CMN\tnote{2} ~(T) & - & - & - & - & 74.1 & - \\
(T+V+A) & {\bf89.88} & 81.75 & 77.73 & 67.32 & 77.6 & \underline{79.1} \\
\hline
bcLSTM$_*$~(T) & 75.29& 79.40& 78.07& 76.53 & 77.7{\scriptsize(1.1)} & 77.3{\scriptsize(1.4)} \\
bcGRU~(T) & 77.20& 80.99& 76.26& 72.50 & 76.9{\scriptsize(1.6)} & 76.7{\scriptsize(1.3)} \\
\hline
HiGRU~(T) & 75.41& {\bf91.64}& 79.79& 70.74 & 80.6{\scriptsize(0.5)} & 79.4{\scriptsize(0.5)} \\
HiGRU-f~(T) & 76.69& 88.91& 80.25& 75.92 & 81.5{\scriptsize(0.7)} & 80.4{\scriptsize(0.5)} \\
HiGRU-sf~(T) &74.78& 89.65& {\bf80.50}& {\bf77.58} & {\bf82.1}{\scriptsize(0.4)} & {\bf80.6}{\scriptsize(0.2)} \\
\hline
\end{tabular}
}
\begin{tablenotes}\footnotesize
\item[1]by \cite{DBLP:conf/acl/PoriaCHMZM17};
\item[2]by \cite{DBLP:conf/naacl/HazarikaPZCMZ18}.
\end{tablenotes}
\end{threeparttable}
\caption{\label{table:IEMOCAP}Experimental results on IEMOCAP. ``(Feat)'' represents the features used in the models, where T, V, and A denote the textual, visual, and audio features, respectively.  The underlined results of bcLSTM and CMN are derived by us accordingly, while ``-'' means the results are unavailable from the original paper.}
\end{table}

% EmotionX
\begin{table*}[t]
\small
\centering
%\captionsetup{belowskip=-1em}
\begin{threeparttable}
\resizebox{\textwidth}{!}{
\begin{tabular}{|l|c|p{0.55cm}p{0.55cm}p{0.55cm}p{0.55cm}|cc|p{0.55cm}p{0.55cm}p{0.55cm}p{0.55cm}|cc|}
\hline
\multirow{2}{*}{\bf Model} 
& \multirow{2}{*}{\bf Train} 
& \multicolumn{6}{c|}{\bf Friends (F)} 
& \multicolumn{6}{c|}{\bf EmotionPush (E)} \\
\cline{3-14}
& & \bf Ang & \bf Joy & \bf Sad & \bf Neu & \bf WA & \bf UWA &
\bf Ang & \bf Joy & \bf Sad & \bf Neu & \bf WA & \bf UWA \\
\hline
\hline
SA-BiLSTM\tnote{1} & F+E & 49.1 & 68.8 & 30.6 & {\bf90.1} & - & 59.6 
& 24.3 & 70.5 & 31.0 & {\bf94.2} & - & 55.0\\
CNN-DCNN\tnote{2} & F+E & 55.3 & 71.1 & 55.3 & 68.3 & - & 62.5 
& 45.9 & 76.0 & 51.7 & 76.3 & - & 62.5 \\
\hline
bcLSTM$_*$ & F(E) & 64.7& 69.6& 48.0& 75.6 & 72.4{\scriptsize(4.2)} & 64.4{\scriptsize(1.6)}
& 32.9 & 69.9 & 47.1 & 78.0 & 74.7{\scriptsize(4.4)}& 57.0{\scriptsize(2.1)}  \\
bcGRU & F(E) & 69.5& 65.4& 52.9& 74.7 & 71.7{\scriptsize(4.7)} & 65.6{\scriptsize(1.2)}
& 33.7& 71.1& 57.2& 76.1 & 73.9{\scriptsize(2.9)} & 59.5{\scriptsize(1.8)}\\
\hline
bcLSTM$_*$ & F+E & 54.5 & 75.6 & 43.4 & 73.0 & 70.5{\scriptsize(4.5)} & 61.6{\scriptsize(1.6)}
& 52.4 & 79.1 & 54.7 & 73.3 & 73.4{\scriptsize(3.8)} & 64.9{\scriptsize(2.1)} \\
bcGRU & F+E & 59.0 & 78.6 & 42.3 & 71.4 & 70.2{\scriptsize(5.1)} & 62.8{\scriptsize(1.4)}
& 49.4 & 74.8 & 61.9 & 72.4 & 72.1{\scriptsize(4.3)} & 64.6{\scriptsize(1.8)} \\
\hline
\hline
HiGRU & F(E) & 66.9& 73.0& 51.8& 77.2 & {\bf74.4}{\scriptsize(1.7)}& 67.2{\scriptsize(0.6)}
& 55.6& 78.1& 57.4& 73.8 & 73.8{\scriptsize(2.0)} & 66.3{\scriptsize(1.7)} \\
HiGRU-f & F(E) & 69.1& 72.1& {\bf60.4}& 72.1 & 71.3{\scriptsize(2.9)} & 68.4{\scriptsize(1.0)}
& 55.9& 78.9& 60.4& 72.4 & 73.0{\scriptsize(2.2)} & 66.9{\scriptsize(1.2)} \\
HiGRU-sf & F(E) & {\bf70.7}& 70.9& 57.7& 76.2 & 74.0{\scriptsize(1.4)} & {\bf68.9}{\scriptsize(1.5)}
& 57.5& 78.4& 64.1& 72.5 & 73.0{\scriptsize(1.6)} & 68.1{\scriptsize(1.2)} \\
\hline
HiGRU & F+E & 55.4 & 81.2 & 51.4 & 64.4 & 65.8{\scriptsize(4.2)} & 63.1{\scriptsize(1.5)}
& 50.8& 76.9& 69.0& 75.7& 75.3{\scriptsize(1.7)} & 68.1{\scriptsize(1.2)} \\
HiGRU-f & F+E & 54.9& 78.3& 55.5& 68.7 & 68.5{\scriptsize(3.0)} & 64.3{\scriptsize(1.2)} 
& {\bf58.3}& 79.1& {\bf69.6} & 70.0 & 71.5{\scriptsize(2.5)} & 69.2{\scriptsize(0.9)} \\
HiGRU-sf & F+E & 56.8& {\bf81.4} & 52.2& 68.7 & 69.0{\scriptsize(2.0)}& 64.8{\scriptsize(1.3)}
& 57.8& {\bf79.3} & 66.3& 77.4 & {\bf77.1}{\scriptsize(1.0)}& {\bf70.2}{\scriptsize(1.1)} \\
\hline
\end{tabular}
}
\begin{tablenotes}\footnotesize
\item[1]by~\cite{DBLP:conf/acl-socialnlp/LuoYC18};
\item[2]by~\cite{DBLP:conf/acl-socialnlp/Khosla18}.
\end{tablenotes}
\end{threeparttable}
\caption{\label{table:EmotionX}Experimental results on Friends and EmotionPush.  In the Train column, F(E) denotes the model is trained on only one training set, Friends or EmotionPush. F+E means the model is trained on the mixed training set while validated and tested individually.}
\end{table*}

% Experimental Results
\subsection{Main Results}

Table~\ref{table:IEMOCAP} and Table~\ref{table:EmotionX} report the average results of 10 trials each on the three datasets, where the standard deviations of WA and UWA are recorded by the subscripts in round brackets.  The results of bcLSTM, CMN, SA-BiLSTM, and CNN-DCNN are copied directly from the original papers for a fair comparison because we follow the same configuration for the corresponding datasets.  From the results, we have the following observations:

\paragraph{(1) Baselines.} 
Our implemented bcLSTM$_*$ and bcGRU, attain comparable performance with the state-of-the-art methods on all three datasets.  

From the results on IEMOCAP in Table~\ref{table:IEMOCAP}, we observe that:
{\bf(a)} By utilizing the textual feature only, bcGRU outperforms bcLSTM and CMN trained on the textual feature significantly, attaining +3.3 and +2.8 gain in terms of WA, respectively. bcLSTM$_*$ performs better than bcGRU, and even beats bcLSTM and CMN with the trimodal features in terms of WA. In terms of UWA, CMN performs better than bcLSTM$_*$ only when it is equipped with multimodal features.
{\bf(b)} By examining the detailed accuracy in each emotion, bcLSTM$_*$ and bcGRU with the textual feature attain much higher accuracy on the neutral emotion than bcLSTM with the only textual feature while maintaining good performance on the other three emotions. The results show that the weighted loss function benefits the training of models.

From the results on Friends and EmotionPush in Table~\ref{table:EmotionX}, we observe that bcLSTM$_*$ and bcGRU trained on the same dataset (F+E) of CNN-DCNN perform better than CNN-DCNN on EmotionPush while attaining comparable performance with CNN-DCNN on Friends.  The results show that by utilizing the contextual information with the weighted loss function, bcLSTM$_*$ and bcGRU can beat the state-of-the-art method.

\paragraph{(2) HiGRUs vs. Baselines.} 
Our proposed HiGRUs outperform the state-of-the-art methods with significant margins on all the datasets.  

From Table~\ref{table:IEMOCAP}, we observe that:
{\bf(a)} CMN with the trimodal features attains the best performance on the anger emotion while our vanilla HiGRU achieves the best performance on the happiness emotion and gains further improvement on sadness and neutral emotions over CMN.  Overall, the vanilla HiGRU achieves at least 8.7\% and 3.8\% improvement over CMN with the textual feature and the trimodal features in terms of WA, respectively. The results, including those of bcLSTM$_*$ and bcGRU, indicate that GRU learns better representations of utterances than CNN in this task. 
{\bf(b)} The two variants, HiGRU-f and HiGRU-sf, can further attain +0.9 and +1.5 improvement over HiGRU in terms of WA and +1.0 and +1.2 improvement over HiGRU in terms of UWA, respectively.  The results demonstrate that the included individual word/utterance-level features and long-range contextual information in HiGRU-f and HiGRU-sf, are indeed capable of boosting the performance of the vanilla HiGRU. 

From Table~\ref{table:EmotionX}, we can see that:
{\bf(a)} In terms of UWA, HiGRU trained and tested on individual sets of Friends and EmotionPush gains at least 7.5\% and 6.0\% improvement over CNN-DCNN, respectively.  Overall, our proposed HiGRU achieves well-balanced performance for the four tested emotions, especially attaining significant better performance on the minority emotions of anger and sadness.  
{\bf(b)} Moreover, HiGRU-f and HiGRU-sf further improve HiGRU +1.2 accuracy and +1.7 accuracy on Friends and +0.6 accuracy and +1.8 accuracy on EmotionPush in terms of UWA, respectively.  The results again demonstrate the superior power of HiGRU-f and HiGRU-sf. 

\paragraph{(3) Mixing Training Sets.} 
By examining the results from the last ten rows in Table~\ref{table:EmotionX}, we conclude that it does not necessarily improve the performance by mixing the two sets of training data.

Though the best performance of SA-BiLSTM and CNN-DCNN is obtained by training on the mixed dataset, the testing results show that our implemented bcLSTM$_*$, bcGRU and our proposed HiGRU models can attain better performance on EmotionPush but yield worse performance on Friends in terms of UWA.

By examining the detailed emotions, we speculate that: EmotionPush is a highly imbalanced dataset with over 60\% of utterances in the neutral emotion.
Introducing EmotionPush into a more balanced dataset, Friends, is equivalent to down-sampling the minority emotions in Friends. This hurts the performance on the minority emotions, anger and sadness. Meanwhile, introducing Friends into EmotionPush corresponds to up-sampling the minority emotions in EmotionPush. The performance of the sadness emotion is significantly boosted and that on the anger emotion is at least unaffected.

\subsection{Discussions}

% Model Size on Friends
\begin{table}[t]
\small
\centering
\begin{threeparttable}
%\resizebox{\columnwidth}{!}{
\begin{tabular}{|c|c|ccc|}
\hline
 \bf $d_1$ & \bf bcGRU & \bf HiGRU & \bf HiGRU-f & \bf HiGRU-sf \\
\hline
\hline
- & 65.6{\scriptsize(1.2)} & - & - & - \\
\hline
300 & - & 67.2{\scriptsize(0.6)} & 68.4{\scriptsize(1.0)} & 68.9{\scriptsize(1.5)} \\
200 & - & 67.6{\scriptsize(2.0)} & {\bf68.9}{\scriptsize(0.9)} & 69.1{\scriptsize(1.3)} \\
150 & - & {\bf67.6}{\scriptsize(1.5)} & 68.5{\scriptsize(1.3)} & 68.9{\scriptsize(1.2)} \\
100 & - & 67.5{\scriptsize(1.7)} & 68.4{\scriptsize(1.3)} & {\bf69.6}{\scriptsize(1.0)} \\
\hline
\end{tabular}
%}
\end{threeparttable}
\caption{\label{table:model_size}Experimental results of UWA on Friends by our proposed models with different scales of utterance encoder.}
\end{table}

\paragraph{Model Size.}
We study how the scale of the utterance encoder affects the performance of our proposed models, especially when our models contain a similar number of parameters as the baseline, say bcGRU.
Such a fair condition can be made between our HiGRU-sf and bcGRU if we set $d_1$ to 150. From the testing results on Friends in Table~\ref{table:model_size}, we can observe that: (1) Under the fair condition, the performance of our HiGRU-sf is not degraded compared to that when $d_1=300$. HiGRU-sf still outperforms bcGRU by a significant margin. (2) Overall, no matter $d_1$ is larger or smaller than 150, HiGRU-sf maintains consistently good performance and the difference between HiGRU-sf and HiGRU-f or HiGRU keeps noticeable. These results further demonstrate the superiority of our proposed models over the baseline bcGRU and the motivation of developing the two variants based on the vanilla HiGRU.

\begin{table}[t!]
\fontsize{10}{11}\selectfont
    \centering
    \resizebox{\columnwidth}{!}{
    \begin{tabular}{|c|p{4cm}|c|c|c|}
    \hline
        \bf Role & \quad\quad\quad \bf Utterance & \bf Truth & \bf bcGRU & \bf HiGRU-sf \\
    \hline
    \hline
        \bf Scene-1 & & & & \\
    \hline
        Phoebe & Okay. Oh but don't tell them Monica's pregnant because they frown on that. & Neu & Neu & Neu \\
    \hline
        Rachel & Okay. & Neu & Neu & Neu \\
    \hline
        Phoebe & Okay. & Neu & Neu & Neu \\
    \hline
    \hline
        \bf Scene-2 \\
    \hline
        Phoebe & Yeah! Sure! Yep! Oh, y'know what? If I heard a shot right now, I'd throw my body on you. & Joy & \colorbox{red!30}{Ang} & Joy \\
    \hline
        Gary & Oh yeah? Well maybe you and I should take a walk through a bad neighborhood. & Other & / & / \\
    \hline
        Phoebe & Okay! & Joy & \colorbox{red!30}{Ang} & Joy \\
    \hline
        Gary & All right. & Neu & Neu & Neu \\
    \hline
    \hline
        \bf Scene-3 & & & & \\
    \hline
        Female & Can I send you, like videos and stuff? What about when they start walking. & Other & / & / \\
    \hline
        Male & Yeah yeah yeah. & Sad & \colorbox{red!30}{Hap} & Sad \\
    \hline
        Male & You you record every second.  You record every second because I want to see it all. Okay? & Hap & Hap & \colorbox{red!30}{Sad} \\
    \hline
        Male & If I don't get to see it now, I get to see it later at least, you know?  You've got to keep it all for me; all right? & Other & / & / \\
    \hline
        Female & Okay. & Sad & \colorbox{red!30}{Neu} & Sad \\
    \hline
    \end{tabular}
    }
    \caption{\label{table:success_cases} ``Okay'' expresses distinct emotions in three different scenes.
    }
\end{table}

\paragraph{Successful Cases.}
We investigate three scenes related to the word ``okay'' that expresses three distinct emotions. The first two scenes come from the testing set of Friends and the third one from that of IEMOCAP. We report the predictions made by bcGUR and our HiGRU-sf, respectively, in Table~\ref{table:success_cases}. 
In {\bf Scene-1}, ``okay'' with period usually exhibits little emotion and both bcGRU and HiGRU-sf correctly classify it as ``Neu". In {\bf Scene-2}, ``okay'' with ``!'' expresses strong emotion.  However, bcGRU misclassifies it to ``Ang" while HiGRU-sf successfully recognizes it as ``Joy".  Actually, the mistake can be traced back to the first utterance of this scene which is also misclassified as ``Ang". This indicates that bcGRU tends to capture the wrong atmosphere within the dialogue. As for {\bf Scene-3}, ``okay'' with period now indicates ``Sad" and is correctly recognized by HiGRU-sf but misclassified as ``Neu" by bcGRU. Note that HiGRU-sf also classifies the third utterance in {\bf Scene-3} as ``Sad" which seems to be conflicting to the ground truth. In fact, our HiGRU-sf captures the blues of this parting situation, where the true label ``Hap" may not be that suitable.
These results show that our HiGRU-sf learns from both each utterance and the context, and can make correct predictions of the emotion of each utterance.

\begin{table}[t!]
\fontsize{10}{11}\selectfont
    \centering
    \resizebox{\columnwidth}{!}{
    \begin{tabular}{|c|p{4cm}|c|c|c|}
    \hline
        \bf Role & \quad\quad\quad \bf Utterance & \bf Truth & \bf bcGRU & \bf HiGRU-sf \\
    \hline
    \hline
        \bf Scene-4 & & & & \\
    \hline
        Ross & Hi. & Neu & Neu & Neu \\
    \hline
        Rachel & Hi. & Neu & Neu & Neu \\
    \hline
        Ross & Guess what? & Neu & Neu & Neu \\
    \hline
        Rachel & What? & Neu & Neu & Neu \\
    \hline
        Ross & They published my paper. & Joy & \colorbox{red!30}{Sad} & \colorbox{red!30}{Neu} \\
    \hline
        Rachel & Oh, really, let me see, let me see. & Joy & \colorbox{red!30}{Neu} & \colorbox{red!30}{Neu} \\
    \hline
        Phoebe & Rach, look!  Oh, hi! Where is my strong Ross Skywalker to come rescue me. There he is. & Other & / & / \\
    \hline
    \hline
        \bf Scene-5 & & & & \\
    \hline
        Speaker-1 & Sorry for keeping you up & Sad & Sad & Sad \\
    \hline
        Speaker-2 & Lol don't be & Joy & Joy & Joy \\
    \hline
        Speaker-2 & I didn't have to get up today & Neu & \colorbox{red!30}{Sad} & \colorbox{red!30}{Sad} \\
    \hline
        Speaker-1 & :p & Joy & Joy & Joy \\
    \hline
        Speaker-2 & It's actually been a really lax day & Joy & \colorbox{red!30}{Neu} & \colorbox{red!30}{Sad} \\
    \hline
    \end{tabular}
    }
    \caption{\label{table:failed_case} Wrong predictions made by both bcGRU and our HiGRU-sf in two scenes.
    }
\end{table}

\paragraph{Failed Cases.}
At last, we show some examples that both bcGRU and our HiGRU-sf fail in recognizing the right emotions in Table~\ref{table:failed_case}, i.e., {\bf Scene-4} from Friends and {\bf Scene-5} from EmotionPush. In {\bf Scene-4}, both bcGRU and HiGRU-sf make wrong predictions for the fifth and the sixth utterances. It should be good news that Ross has his paper published and Rachel is glad to see related reports about it. However, the transcripts do not reveal very strong emotions compared to what the characters might act in the TV show. This kind of scenes may be addressed by incorporating some other features like audio and video. As for {\bf Scene-5}, the third and the fifth utterances are classified into wrong emotions. Notice that the emotions indicated from the two utterances are very subtle even for humans. The Speaker-2 did not plan to get up today, but Speaker-1 kept him/her up and it ended up with a really lax day. So, the Speaker-2 feels joyful now. This indicates that even taking into the context into account, the models' capability of understanding subtle emotions is still limited and more exploration is required.

\section{Conclusion}

We propose a hierarchical Gated Recurrent Unit (HiGRU) framework to tackle the utterance-level emotion recognition in dialogue systems, where the individual utterance embeddings are learned by the lower-level GRU and the contexts of utterances are captured by the upper-level GRU.  We promote the HiGRU framework to two variants, HiGRU-f, and HiGRU-sf, and effectively capture the word/utterance-level inputs and the long-range contextual information, respectively.  Experimental results demonstrate that our proposed HiGRU models can well handle the data imbalance issue and sufficiently capture the available text information, yielding significant performance boosting on all three tested datasets. In the future, we plan to explore semi-supervised learning methods to address the problem of data scarcity in this task.

\section*{Acknowledgments}
This work is supported by the Research Grants Council of the Hong Kong Special Administrative Region, China (No.~CUHK 14208815 and No.~CUHK 14210717 of the General Research Fund, and Project No.~UGC/IDS14/16), and Meitu (No. 7010445). We thank the three anonymous reviewers for the insightful suggestions on various aspects of this work.

% references
%\newpage
\bibliography{naaclhlt2019}
\bibliographystyle{acl_natbib}

\appendix

\end{document}